\documentclass[conference]{IEEEtran}
\IEEEoverridecommandlockouts
\usepackage{cite}
\usepackage{amsmath,amssymb,amsfonts}
\usepackage{algorithmic}
\usepackage{graphicx}
\usepackage{textcomp}
\usepackage{xcolor}
\usepackage{gensymb}
\usepackage{todonotes}

\usepackage{subfig}

\def\BibTeX{{\rm B\kern-.05em{\sc i\kern-.025em b}\kern-.08em
    T\kern-.1667em\lower.7ex\hbox{E}\kern-.125emX}}
    
\definecolor{light-gray}{gray}{0.90}

\newcommand{\game}{Rinascimento}
\newcommand{\splendor}{Splendor\texttrademark}
\newcommand{\gameR}{$\mathcal{R}$}
\newcommand{\gameS}{$\mathcal{S}$}
\newcommand{\FPS}{4P$\mathcal{S}$}
\newcommand{\param}[1]{\textit{#1}}
\newcommand{\n}{n\textdegree~}

\begin{document}

\title{\game:  Optimising Statistical Forward Planning Agents for Playing Splendor}

\author{\IEEEauthorblockN{Ivan Bravi, Diego Perez-Liebana and Simon M. Lucas}
\IEEEauthorblockA{\textit{School of Electronic Engineering and Computer Science}\\
\textit{Queen Mary University of London}\\
London, United Kingdom \\
\{i.bravi, diego.perez, simon.lucas\}@qmul.ac.uk}
\and
\IEEEauthorblockN{Jialin Liu}
\IEEEauthorblockA{
\textit{Shenzhen Key Laboratory of Computational Intelligence}\\
\textit{Department of Computer Science and Engineering}\\
\textit{Southern University of Science and Technology}\\
Shenzhen, China\\
liujl@sustech.edu.cn}
}
\maketitle
\begin{abstract}
Game-based benchmarks have been playing an essential role in the development of Artificial Intelligence (AI) techniques. Providing diverse challenges is crucial to push research toward innovation and understanding in modern techniques. \game~ provides a parameterised partially-observable multiplayer card-based board game, these parameters can easily modify the rules, objectives and items in the game.
We describe the framework in all its features and the game-playing challenge providing baseline game-playing AIs and analysis of their skills.
We reserve to agents' hyper-parameter tuning a central role in the experiments highlighting how it can heavily influence the performace.
The base-line agents cointain several additional contribution to Statistical Forward Planning algorithms.
\end{abstract}




\section{Introduction}

The bond between Artificial Intelligence (AI) and games goes back to the origins of AI itself. AI can be used in games in a multitude different ways: to procedurally generate content (PCG), to control non-player characters or opponents, to balance their difficulty, but also to generate complete games (an extensive collection of AI applications in games can be found in \cite{yannakakis2018artificial}).
In academia, advancements in AI are fostered through periodic competitions that help benchmarking the level of new AI techniques. Competitions are based on open-source frameworks which often propose different tracks for different tasks e.g. 1 or 2-player game-playing or level generation.
In the case of game-playing, there are mainly two branches of competitions targeting learning-based algorithms and search-based algorithms. Game-playing learning algorithms usually require big computational budgets to be trained and are generally able to play a single game once trained \cite{mnih2013playing}. On the other hand search-based algorithms, also known as planning algorithms, use a Forward Model (FM) of the game to simulate possible future states.
Several frameworks encourage advancements on Artificial General Intelligence (AGI) benchmarking such algorithms on a wide selection of different games such as General Video Game AI (GVGAI) \cite{perez2018general}.
However it's worth noticing how often with little modifications it is possible to drastically modify games. Card games based on the 52 card decks (e.g. Bridge and Poker) are clear examples.

\section{Motivation}

    In this paper we will present \game~(\gameR) a game framework based on the popular board game \splendor~(\gameS) published by Space Cowboys in 2014 and designed by Marc Andr\'{e}. \gameS~is a turn-based multiplayer (from 2 to 4 players) board game where the players race against each other to obtain the most wealth and prestige. The games revolves around randomly shuffled decks of cards, token stacks and randomly-selected noble tiles (for more details see Section \ref{sec:splendor})
    
    The game engine is implemented in a parameterised way, every rule controlling the mechanics of the game can be tweaked by changing its parameters. This unlocks many applications from the perspective of AI applied to game design. A similar approach was taken in \cite{isaksen2015discovering} where the authors altered the parameters in the popular smartphone game Flappy Bird. They showed how the game design space can be searched for unique variants that suit completely different skill sets.
    
    The parameters can actually influence the type of content used during the game: cards, tokens, noble tiles. This makes this game also particularly interesting to integrate PCG and game-playing in a single framework.
    
    This game provides new challenges for game AI since it is a highly-stochastic partially-observable multi-player game.
    
    We decided to develop the AIs for this framework giving much relevance to hyper-parameter tuning. This is an aspect that has been often overlooked in the past. In \cite{lucas2019efficient} is shown how a correct tuning can make-or-break agent's performance. Thinking about game-playing AI in terms of tunable algorithms will get us closer to measure their true potential, hand-picked parameters can severely limit their performance.

\section{Background}
        \subsection{Game AI frameworks}
        
        A number of game-based frameworks have been designed and implemented for different research purposes. The most classic ones include, but not limited to: GVGAI~\cite{perez2018general}, microRTS Framework~\cite{ontanon2013combinatorial}, Mario AI Framework~\cite{shaker20112010}, AI Bird Framework (AIRBIRDS) ~\cite{renz2015aibirds}. Each one has one or more relative competitions periodically run at major Game AI conferences. This trend started mainly around game-playing tasks, to then expand to PCG: GVGAI, AIRBIRDS and Mario AI have PCG tracks where content is produced and evaluated by humans.
        
        When the task is game-playing AI agents are usually provided with a bounded action space either by enumerating the actions (e.g: GVGAI and Mario AI), or through an explicit a-priory knowledge of the action space (e.g.: microRTS).
        
        In the past three years, the frameworks for Text-Based Adventure AI Competition~\cite{atkinson2019text}, Generative Design in Minecraft Competition \cite{salge2018generative} and (MARLO) \cite{perez2019multi} have been released. These frameworks highlight the need of more complex scenarios to test AIs shifting the attention to 1-player and 2-player to multi-player games, from 2D to 3D PCG, and from fully observable to partially-observable game states.
        The Hearthstone AI Competition~\cite{dockhornheartfstone} has two tracks: Pre-made Deck Playing and User Created Deck Playing. The second is particular interesting because combines together high-level PCG and game-playing in a single challenge.
        
        Such frameworks can also be used to approach more game-design-related problems. In \cite{kunanusont2018modeling} the authors optimised parameters that govern the rules of several GVGAI games to modify the player's experience.
    
    \subsection{Game-playing AI}
    
        General frameworks based on board games can be implemented with extremely fast forward models, particularly suitable for Statistical Forward Planning (SFP) game-playing AIs.
        SFP methods, such the ones later defined, can provide overall good performance in many different scenarios, as shown by their results on the planning tracks of GVGAI \cite{perez2018general}.
    
        \subsubsection{Monte-Carlo Tree Search}
            Monte-Carlo Tree Search (MCTS)~\cite{browne2012survey} has been the state-of-the-art for planning in games in the last years, being successfully applied to both deterministic and stochastic games with perfect or partial information~\cite{cowling2012information}. In \cite{browne2012survey}, Browne et al. review the advances and usages of MCTS till 2012. 
            \textit{MoGo}~\cite{gelly2005modification}, the first computer Go program using Upper Confidence Tree (UCT), reduced the branching factor and the length of random simulations using a pattern group pruning technique and a zone pruning technique, respectively. Thus, instead of considering the whole board, only a sub-group of patterns or a sub-zone of the board is considered~\cite{gelly2005modification}.
            R. Coulom applied progressive widening to MCTS in his Go program \textit{CRAZY STONE} to perform a local search~\cite{coulom2007computing}. Chaslot et al.~\cite{chaslot2007progressive} proposed progressive bias and progressive unpruning to enhance their Go program, \textit{MANGO}.
        
        \subsubsection{Rolling Horizon Evolutionary Algorithm}
            Rolling Horizon Evolutionary Algorithms (RHEAs)~\cite{perez2013rolling} model action sequences of a fix horizon $t_p$ as a population of integer vectors at time $t$. Only the first $t_o$ action(s) ($1 \leq t_o < t_p$) of the approximate optimal action sequence are applied, then at time $(t + t_o)$, a new population is initialised and evolved for the next $t_p$ time steps with the updated environment. This procedure is also called receding horizon control or model predictive control.
            RHEAs was firstly applied to Physical Travelling Salesman Problems (PTSPs) in 2013~\cite{perez2013rolling}, then quickly became popular and achieved competitive results with MCTS in GVGAI. 
            The impact of the planning horizon and population size of RHEAs has been studied in \cite{gaina2017analysis}.
            Gaina et al.~\cite{gaina2017rolling,gaina2017population} designed several enhancement techniques for RHEAs in general video game playing, such as shifted-buffer and population seeding.

    \subsection{Automatic Algorithm Configuration}
    
         Algorithms usually are dependent on few parameters that affect how they function. These can be ordinal, categorical or numerical. Classic examples can be: the learning rate in a learning algorithm, the mutation operator used in RHEA or also terms of equations such those in the tree policy in MCTS (UCB1). Being able to explore different configurations of the algorithm can grant significant improvements in the global performance of the agent.
        For this purpose, different automatic algorithm configuration frameworks have been proposed, including model-based approaches and model-free approaches, such as SMAC~\cite{hutter2011sequential} and the recently proposed  NTBEA~\cite{lucas2018n,lucas2019efficient}.
        Bravi et al.~\cite{bravi2017evolving} evolved UCB alternatives for general video game playing using genetic programming. Sironi et al.~\cite{sironi2018selfggp} compared NTBEA to CMA-ES in evolving MCTS in real-time for general game playing.
        NTBEA has also shown good results on both agent\cite{lucas2019efficient} and game tuning \cite{lucas2018n}.

\section{\splendor} \label{sec:splendor}
\begin{figure}[!t]
    \centering
    \includegraphics[width=\columnwidth]{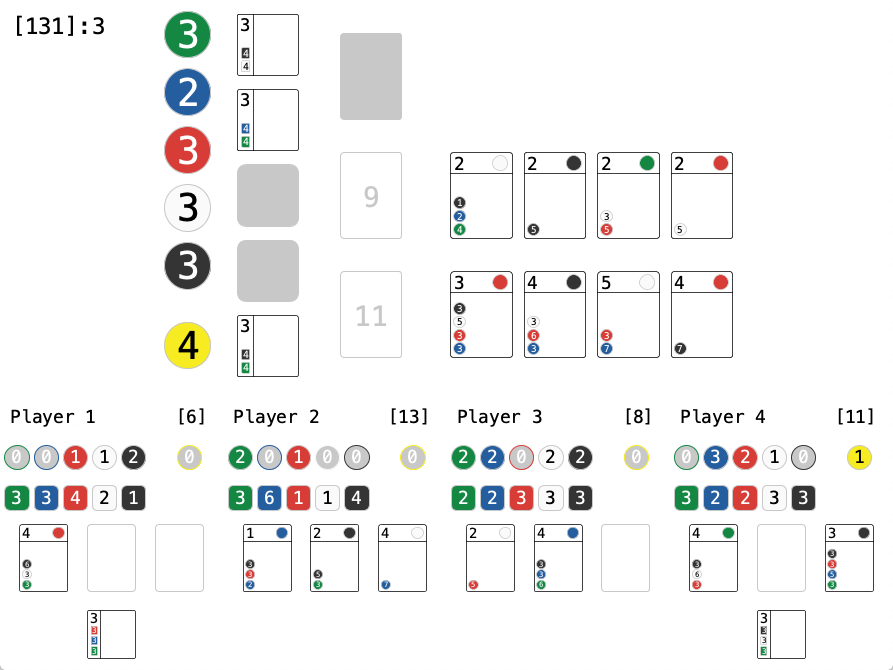}
    \caption{\label{fig:board} Game state represented in the framework's UI.}
    \end{figure}

    In the following we are going do describe the main elements in the game \gameS.
    The game comes with three types of items: tokens, development cards and noble tiles. Figure \ref{fig:board} shows a typical setup of the game.
    There are 2 types of tokens:
    \begin{itemize}
        \item common token: a token has one of five suits (emerald, diamond, sapphire, onyx and ruby) and there is a total of seven tokens for each suite;
        \item joker token: its suit is gold, it can be used as any common token, there is a total of five tokens.
    \end{itemize}
    Cards are characterised by three bits of information:
    \begin{itemize}
        \item bonus: the suit (same as common tokens) of the card;
        \item price: amount of tokens required for each suit to buy it;
        \item value: amount of prestige points.
    \end{itemize}
    Cards are divided in three decks: level 1 (40 cards), level 2 (30 cards) and level 3 (20 cards). As the level increases so do cost and value of the cards in the deck. 
    In the game there are nine noble tiles, each noble is characterised by a value (prestige points) and by an amount of bonuses.

    \subsection{The rules}

        The game setup varies with the number of players later denoted as $p$. The decks and the nobles are shuffled, on the table are placed
        \begin{itemize}
            \item $p+1$ randomly-picked nobles;
            \item $p+2$ common tokens for every suit;
            \item all joker tokens;
            \item four card for each deck.
        \end{itemize}

        From this state the game is played in turns during which the player can play one of the following actions: pick tokens, reserve a card, buy a card.

        Players can have in their hand a maximum of ten tokens (regardless of suit) and three reserved cards. If after an action the player has more than ten tokens, they can give back tokens of any suit until the tokens count is down to the maximum allowed.
        A player can pick from the table either up to three common tokens of a different suit (\textit{pick different}) or two of the same suit (\textit{pick same}). The stack of the token type chosen for a \textit{pick same} action must have at least 4 tokens.
        Players can reserve cards either from the ones face-up on the table (\textit{reserve table}) or drawing the first one from one of the decks (\textit{reserve deck}). Reserving a card grants the player a joker token if there are any left on the table. If the card is reserved from one of the decks, the player can look at it but then the card is kept face-down until purchased.
        Cards can be bought if they are on the table face-up (\textit{buy table}) or between the player's own reserved cards (\textit{buy reserved}). To buy a card the player pays the amount of tokens specified on the card: paying means putting back the coins on the table. Players can get a discount on the price based on the cards they own, each card grants a discount on the token suit specified by the card's bonus.

        These were \textit{active actions}, the game also comes with \textit{passive actions} which are actions triggered by the game at the end of each turn.
        \gameS~has only one: whenever a player has the exact amount of bonuses specified by a noble tile, that player automatically acquires the noble and its prestige points, however it is possible to gain only one noble per turn.

        Each player's prestige points are calculated summing the points of the cards they bought and the nobles they have got.
        Once one of the players reaches 15 prestige points the round is completed and once finished the game is over.
        When the game over condition is reached the player with the most prestige points wins. If two or more players have the same amount of points, the player with less cards wins. If several players have the same points and cards they all win.

    \subsection{Game dynamics and features}

        This game is particularly relevant for game AI because of the balance of simple and complex elements. It provides complex challenges but in a scenario that can be clearly analysed. Simple elements:
        \begin{itemize}
            \item game state representation: information is simple, everything revolves around type and amount of token/bonus, plus the structure of this information is very precise, as shown in the first paragraphs of Section \ref{sec:splendor}.
            \item actions have immediate effects on the game state; 
            \item atomic and simple events can be clearly identified, e.g.: taking or giving back tokens, getting a card, etc;
        \end{itemize}
        Complex elements:
        \begin{itemize}
            \item long-term implications of early-game actions;
            \item games are limited in time;
            \item there are elements of partial observability;
            \item and, last but not least, it's a multi-player game, opponent modelling can be beneficial for your own strategy.
        \end{itemize}

        The gameplay arising from the simple rules is quite complex and require thorough planning and prediction of possible opponents strategies. Moreover the relationships between game elements are quite intricate and they result in a gameplay where every action matters and has an influence till the end of the game.
        Here we highlight the complexity before mentioned through few examples:
        \begin{itemize}
            \item getting hold of a card with a rare bonus (due to shuffling) in the early game can be crucial for the final outcome;
            \item reserving a card can stop one of the players from winning and delaying the game-over enough to come up with a winning strategy;
            \item coin scarcity limits opponents' ability of buying cards;
        \end{itemize}

\section{The game parameters}

    Diving into \gameS's rules we can easily recognise all the elements that can be parameterised.
    Implementing \game's game engine using the parameters of the games rather than explicit values allows us to reason on more abstract terms on what are the abstract game mechanics. More importantly it allows us to implement an engine that is able to play not only the base \gameS~game but the entirety of \gameS-like games.

    \begin{table}[!t]
        \centering
        \caption{Parameters extracted from the game's setup. The ones marked with a star require PCG.}
        \begin{tabular}{l|c|c}
        \textbf{Description}     & \textbf{Symbol}  & \textbf{Default}   \\
        \hline
        \n players              &\param{P}          & 4         \\
        token types*             &\param{nTT}        & 5         \\
        \n joker token          &\param{nJT}        & 5         \\
        \n decks*                &\param{D}          & 3         \\
        \n face-up cards        &\param{FUC}        & 4         \\
        \n extra noble*          &\param{EN}         & 1       
        \end{tabular}
    \label{par:setup}
    \end{table}

    \begin{table}[!t]
        \centering
        \caption{Parameters extracted from the game's rules.}
        \begin{tabular}{l|c|c}
        \hline
        \textbf{Description}        & \textbf{Symbol}   & \textbf{Default}   \\
        \hline
        max \n tokens per player    & \param{maxT}      & 10        \\
        max \n reserved cards       & \param{maxRC}     & 3         \\
        end-game prestige points    & \param{PP}        & 15        
        \end{tabular}
    \label{par:rules}
    \end{table}

    \begin{table}[!t]
        \centering
        \caption{Parameters from the game's actions rules.}
        \begin{tabular}{l|c|c}
        \hline
        \textbf{Description}                                    & 1\textbf{Symbol}   & \textbf{Default}   \\
        \hline
        \n different token types in \textit{pick different}     &\param{nTTPD}      & 3         \\
        \n tokens per type in \textit{pick different}           &\param{nTPD}       & 1         \\
        \n tokens in \textit{pick same}                         &\param{nTPS}       & 2         \\
        min \n available tokens in \textit{pick same}           &\param{minTPS}     & 4            
        \end{tabular}
    \label{par:actions}
    \end{table}

    In Tables \ref{par:setup}, \ref{par:rules} and \ref{par:actions} are listed all the parameters, for each we specify its acronym and the value for 4 players \gameS~(later addressed as \FPS). Together the form a total of 14 parameters, they can be seen as a 14-dimensional vector of integer values [\param{P}, \param{nTT}, \param{nJT}, \param{D}, \param{FUC}, \param{EN}, \param{maxT}, \param{maxRC}, \param{PP}, \param{nTTPD}, \param{nTPD}, \param{nTPS}, \param{minTPS}]. 
    The game however relies on some content (cards and nobles) which is dependent on the parameter. Thus the game engine will need to a procedural content generation components that take care of generating cards and noble tiles, this feature however is left for future expansion.

\section{The framework}
    The \game~framework (\gameR~in short)\footnotemark~is engineered to be extremely flexible and customisable. It is not limited to a parametric implementation of the game rules, in fact, new actions can be implemented (both active or passive).
    \footnotetext{link to repository with documentation upon publication}
    The code base was developed in Java due to its low barriers to new developers (encouraging its adoption), high-efficiency of its garbage collection (lifting the burden of explicit memory management) and its innate cross-platform compatibility.

    There are four main components in its design: Player, State, Engine, Action;
    The State is the object that encodes the game state, more specifically: stacks of tokens, cards and nobles on the table, decks and, finally, tokens and cards in the players' hands.
    The Action object represents, as the name suggests, an action from a specific player encoding internally all the relative information .
    The Player is the entity that is responsible to provide an Action during its turn, it can use the State to retrieve random actions that could be performed by a specified player.
    The Engine is the object responsible for:
    \begin{itemize}
        \item owning the action space used by the agent in the game;
        \item triggering the passive rules;
        \item calling the players for their next action thus managing the turns in the game;
        \item checking end-game conditions (stalemate (SM) or game-over);
     \end{itemize} 

\begin{figure}[!t]
    \centering
    \includegraphics[width=0.9\columnwidth]{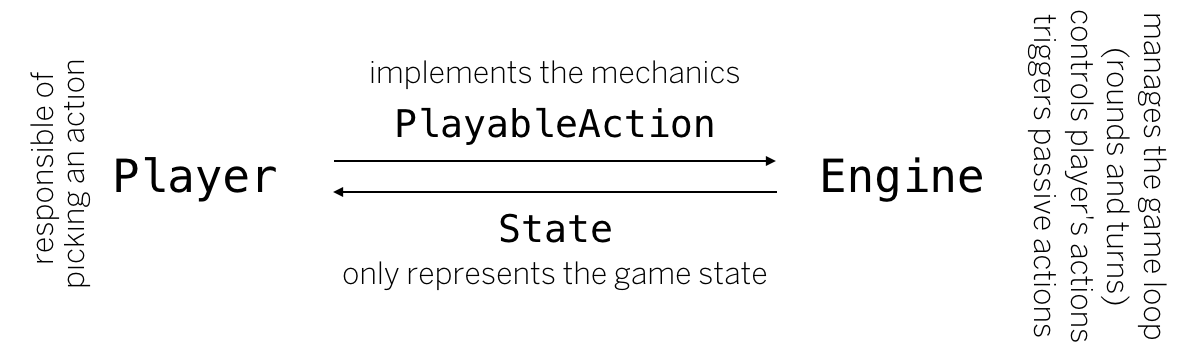}
    \caption{\label{fig:arch} Interactions between core components in \gameR.}
    \end{figure}

    Figure \ref{fig:arch} shows the interactions and the duties of the components.

            This subsection highlights the main features that differentiate the framework from other existing frameworks used for Game AI.

        \subsubsection{Automatic Player's Budget Management}
            In most frameworks used for game-playing AI benchmarking the equaliser is some kind of budget give to the AI. The budget is usually either an amount of forward model calls or CPU time, although other forms of budget could be adopted, possibly more sophisticated such as a combination of the two.
            The framework provides to the player a game state coupled with a resource checker that can be queried on the amount of \textit{budget} left. Once the budget is over the player won't be able to use the forward model anymore, if the player tries to use an expired budget an exception is thrown to further warn the player. Currently the framework implements a budget based on FM calls.

            Up to date, most game AI frameworks have not paid much attention to easing the use of opponents models to better shape your own strategy. When adopting player modelling we basically allot some of the budget to predict opponents' decisions. In \gameR~this is very easy, in fact, thanks to this budget management system it is possible to split a portion of your own budget, e.g. a percentage of the original budget, and give it to the opponent model. In case just a portion of the budget is used, the original budget is depleted only by that portion. In other frameworks this has to be handled explicitly requiring ad-hoc code depending on the kind of budget, but in \gameR~this is seamless.

        \subsubsection{Action Space and Forward Model}
            \gameR's Action Space (AS) is modular implemented through independent and atomic PlayableAction (PA) objects, this architecture allows to separate the different mechanics in the game making easy to implement new mechanics. Thus the AS is defined withing the Engine as a collection of PAs that can operate on any State whether it's the real game state or a simulated one. The Forward Model (FM) can be in fact the same collection of PAs in the Engine or a separate one to provide a different (or imperfect) FM to the player.
            In addition also passive rules can be added.

        \subsubsection{Random Action Generator}
            One of the core challenges in designing the framework was on providing a universal interface for an AGI player. Frameworks usually enumerate the actions available and that's quite easy for actions like buying or reserving cards. However enumerating pick actions is quite complex, e.g. when a player also need to give back some tokens. In that case a nested combinatorial problem needs to be solved facing heavy computation especially keeping in mind the parametric nature of the game i.e. combinations explode with increasing the token types. 
            
            We decided to provide an interface to a Random Action Generator (RAG). This can be seen as a tool that samples the action space hiding its real complexity. 
            Since the action space is completely customisable, and the PA directly implements the mechanics, introducing a new action simply means providing to the engine a new RAG.
            A peculiar feature of a RAG is that it can generate a random action for a specific player given its id based on a seed. The seed is used by the RAG to generate the action, using it the agent can influence how the action space is sampled. By contract the RAG returns a \textit{null} when it's not possible to perform any action of such kind.

        \subsubsection{Stalemate detection}
            Similarly to Chess, a stalemate condition happens when players can't play a legal action during their turn.
            This feature is implemented through \textit{exact} RAGs: when none of the generators is able to produce a non-\textit{null} action, a \textit{StalemateException} is thrown.
            Another danger is ending up in cyclic game states: in the game it is possible to take actions that don't change the game state, i.e. a pick action when the player has already \param{maxT} tokens and puts back the same tokens that were picked. Such condition is avoided limiting the number of ticks per game.

            
            Games can be run with or without visual, a very simple UI is provided with the framework and it adapts seamlessly when varying game parameters.


        The framework has a remarkably fast Forward Model, running \FPS~with random actions we registered the following stats\footnotemark: speed 1.74 M states/s, average game duration 0.44 ms, 14.1\% stalemate rate.


        \footnotetext{Run on Intel(R) Core(TM) i7-3615QM (2012) CPU @ 2.30GHz, 8GB 1600 MHz DDR3 RAM.}

\section{The AI Agents}
    We have implemented several AI agents: two basic policies to provide a baseline and three more sophisticated based on state of the art algorithms in video game playing.
    The latter agents have been implemented keeping in mind the highly parameterised nature and flexibility of their algorithms. In fact their hyper-parameters can be tuned using an optimisation algorithm. We have implemented two different versions of Rolling Horizon Evolutionary Algorithm (RHEA) agent and one Monte-Carlo Tree Search (MCTS) agent.
    All the advanced agents can use models for the opponents, the specific model is defined through an hyper-parameter $om$, the options are do-nothing agent (0), random agent (1) and one-step look ahead (2). A budget can be provided to these models and it's controlled by the $omsb$ hyper-parameter. Since $om$ and $ombs$ are common to all the agents they are omitted.
    Moreover all the advanced algorithms rely on a heuristic to search the action space, it is possible to come up with several ones for \gameS, we used one that comes naturally from the rules: the number player's prestige points. With such heuristic the algorithms were setup as for maximisation problems.

    \subsection{Basic Agents}
        The Random Agent (RND) is a player that performs, as the name suggests, random actions. It simply returns the first random action generated by the game state.
        The One-Step Look Ahead (OSLA) agent instead keeps sampling random actions keeping track of the best (according to a heuristic) action until the budget is over.

    \subsection{Branching Mutation Rolling Horizon Agent} 
        The Branching Mutation Rolling Horizon (BMRH) agent is implemented along the lines of \cite{perez2013rolling} this agent evolves sequences of explicit actions, by explicit we mean an actual Action Java Object.
        The main new contribution to this agent is the usage of a different kind of mutation operator: branching mutation. In standard RHEA, mutating an action sequence simply means swapping an action id with a new one. Such id is essentially an index number that is unequivocally mapped to an action. This works perfectly when the action space is fixed or it can be easily enumerated but this is not the case in \gameR. In fact, the AS highly depends on the current state.

        To initialise the first sequence it is sufficient to (1) request a random action, (2) execute it and keep repeating (1) and (2) until the end of the sequence. However when it comes to mutating an action it is necessary to roll the current game state through the sequence up to the action we want to mutate in order to get legal (and meaningful) random actions to substitute it with.
        This essentially means to potentially follow the same path through game states, up to stochasticity.

        \begin{figure}[!t]
        \centering
        \includegraphics[width=0.7\columnwidth]{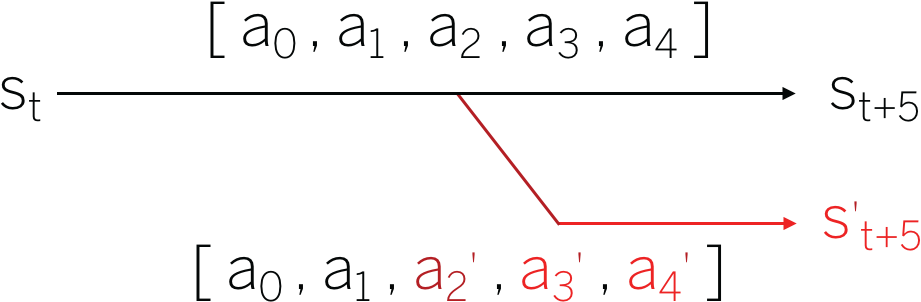}
        \caption{\label{fig:branching} shows an example of branching mutation on a sequence of length 5. $a_2$, thus the selected mutation point, then $a_{3-4}$ are the following random actions rolled.}
        \end{figure}

        The branching mutation operator picks an index in the sequence and from there on it starts mutating the remaining actions while rolling the state.
        Figure \ref{fig:branching} shows the core idea behind the branching mutation. The mutation point can be selected using three different distributions: uniformly across the sequence, with exponential decay from starting probability $dcy$, following a gaussian distribution with mean $\mu$ and standard deviation $\sigma$. $dcy$, $\mu$ and $\sigma$ are hyper-parameters of the agents, a full list can be found in Table \ref{tab:bmrh}.
        Other than branching mutation this agent can be tuned to use no mutation at all or a complete different sequence. This agent requires what we could call an online mutation: evaluation is done at the same time as the mutation since the state has to be rolled.

        \begin{table}[!t]
            \centering
            \caption{\label{tab:bmrh} Hyper-parameters of the BMRH agent.}
            \begin{tabular}{l|c|l}
            \hline
            \textbf{Symbol}     &\textbf{Type}  & \textbf{Description}\\
            \hline
            $l$                 &integer        & sequence length \\
            $n$                 &integer        & sequences evaluated\\
            $usb$               &boolean        & if it uses shift buffer\\
            $mo$                &boolean        & if it has to mutate once\\
            $ms$                &integer        & mutation type\\
            $dcy$               &double         & probability of exponential decay\\
            $\mu$               &double         & mean of the gaussian mutation point\\
            $\sigma$            &double         & std dev of the gaussian mutation point
            \end{tabular}
        \end{table}

    \subsection{Seeding Rolling Horizon Agent}
        The Seeding Rolling Horizon (SRH) agent is another variation of RHEA, it exploits one of the feature of \gameR~: the possibility to provide a seed to the RAG. The sequence evolved is made of \textit{long} seeds which are going to be used to generate deterministically the action sequence to perform. Using such action encoding it's possible to mutate and evaluate the sequences separately, offline, as opposed to BMRH. A similar approach was used to optimise stochastic agent's performance in \cite{liu2016fast}, but in this case seeds are used to bias the action generators to provide better actions. Using seeds might not be as robust as dealing with actual action plans, in fact, the search space is theoretically infinite although practically limited by Java's \textit{long} precision, thus much harder to search.

        Since we decoupled mutation and evaluation, mutation operators more similar to the standard RHEA can be used.

        \begin{table}[!t]
            \centering
            \caption{\label{tab:srh} Hyper-parameters of the SRH agent.}
            \begin{tabular}{l|c|l}
            \hline
            \textbf{Symbol}     &\textbf{Type}  & \textbf{Description}\\
            \hline
            $l$                 &integer        & sequence length \\
            $n$                 &integer        & sequences evaluated\\
            $usb$               &boolean        & if it uses shift buffer\\
            $mo$                &boolean        & if it has to mutate once\\
            $mr$                &double         & mutation probability\\
            \end{tabular}
        \end{table}

    \subsection{Monte Carlo Tree Search Agent} 
        The main feature of this implementation is its ability of dealing with the unknown size of the action space similarly to progressive widening. In the following we describe our implementation broken down in the classic 4 steps:
        \begin{itemize}
            \item 1)Selection: the algorithm travels from the root towards the leaves. Every step the selected node's action is performed together with the opponents' according to their model. Selection is done as follows:
                \begin{itemize}
                    \item if current node is terminal: jumps to 4;
                    \item else if the node wasn't expanded it jumps to 2;
                    \item else with probability $ep$: jump to 2;
                    \item else pick children with highest UCB, jump to 1;
                \end{itemize}
            \item 2)Expansion: the algorithm samples the actions space $ps$ times adding a maximum of $ps$ nodes to the current node, one for each unique action sampled. One newly expanded node becomes the current and it proceeds to 3;
            \item 3)Rollout: from the current node a random rollout is carried out until $d$ depth is reached then goes to 4;
            \item 4)Backpropagation: the reward $r$ is backpropagated up the tree. $r$ is the heuristic delta from game state reached after the rollout and the present game state. The statistics in the nodes traversed are updated.
        \end{itemize}

        Once the algorithm has consumed the budget available it will return an action using either one of three recommendations based on $rt$: max child (0), robust child (1) or secure child (2).
        For MCTS's hyper-parameters see Table \ref{tab:mcts}.

        \begin{table}[!t]
            \centering
            \caption{\label{tab:mcts} Hyper-parameters of the MCTS agent.}
            \begin{tabular}{l|c|l}
            \hline
            \textbf{Symbol}     &\textbf{Type}  & \textbf{Description}\\
            \hline
            $d$                 &integer        & max depth reached by the tree or the rollout \\
            $c$                 &double         & exploration constant of UCB\\
            $e$                 &double         & $\epsilon$ of UCB\\
            $ep$                &double         & probability of further expanding the node \\
            $ps$                &integer        & number of actions sampled during expansion\\
            $rt$                &integer        & recommendation type\\
            \end{tabular}
        \end{table}

\section{Methods}
    In this section we describe the experiments we have carried out and their objective.
    All the experiments are based on the \FPS version of the game, the agent i given a 1000 action-simulations budget per tick. 
    We have first carried out a \textit{preliminary experiment} (1000 games between 4 RND agents) to understand some features of the game from an AI perspective: average length and probability of stalemate.
    
    Then, in order to test the abilities of our BMRH, SRH and MCTS agents, we have run a \textit{grid search} using the parameters in Table \ref{tab:hpspace}. Each configuration of the algorithms was tested over 1000 \FPS~games against 3 other OSLA agents. The values for $om$ and $omsb$ are respectively $\{0,1,2\}$ and $\{0.005,0.01,0.02,0.05\}$
    The parameters were hand-picked using the authors' knowledge of the algorithms and the domain.
    Our focus with these experiments is to highlight the sensibility of the algorithms to their hyper-parameters.
    
    \begin{table*}[!t]
            \centering
            \caption{\label{tab:hpspace}Hyper-parameters spaces (total size of the space between parenthesis). In bold, the best parameter value found.}
            \begin{tabular}{l|l|l|l|l|l}
            \hline
            \multicolumn{2}{c|}{BMRH} & \multicolumn{2}{c|}{SRH} & \multicolumn{2}{c}{MCTS}\\
            \hline
            \textbf{Parameter}  &\textbf{Values (207,360)} &\textbf{Parameter}  &\textbf{Values (28,800)}&\textbf{Parameter}  &\textbf{Values (32,400)}\\
            \hline
            $l$                 & $\{1,\mathbf{2},3,5,10,20\}$           &$l$     & $\{1,\mathbf{2},3,5,10,20\}$               &$d$        &$\{\mathbf{2},3,6,11,21\}$     \\        
            $n$                 & $\{20,50,100,\mathbf{200}\}$           &$n$     & $\{0,1,5,10,20,50,100,\mathbf{200}\}$      &$c$        &$\{\mathbf{0.0},1.41,4.0,9.0,15.0,20.0\}$     \\    
            $usb$               & $\{false,\mathbf{true}\}$              &$usb$   & $\{false,\mathbf{true}\}$                  &$e$        &$\{\mathbf{1.0E-6}\}$     \\   
            $mo$                & $\{false,\mathbf{true}\}$              &$mo$    & $\{\mathbf{false},true\}$                  &$ep$       &$\{0.1,0.2,0.3,\mathbf{0.4}\}$     \\ 
            $ms$                & $\{0,\mathbf{1},2\}$                   &$mr$    & $\{0.01,0.05,0.1,0.2,0.3,0.5,0.7,0.8,\mathbf{0.9},1.0\}$ &$ps$  &$\{\mathbf{1},3,5,10,15\}$     \\
            $dcy$               & $\{0.5,0.7,\mathbf{0.8},0.9\}$         &        &                                 &$ombs$     &$\{0.01,\mathbf{0.05},0.1\}$     \\
            $\mu$               & $\{0.0,\mathbf{0.1},0.3,0.5,0.75\}$    &        &                                 &           &     \\
            $\sigma$            & $\{\mathbf{0.5},1.0,2.0\}$             &        &                                 &           &     
            \end{tabular}
        \end{table*}

    However, even if grid search gives a broader representation of the hyper-parameter space, it is often extremely expensive to run. Thus we designed another experiment to check the feasibility of using a hyper-parameter tuner to reduce the computational used. Multiple experiments are run to see how the agents can be tuned appropriately varying the NTBEA's budget. The true fitness, i.e. the win ratio, is measured over 1000 games with the suggested configuration. Each experiment (fixed budget and agent type) is run 100 times and the results are shown in box plots to compare the outcomes. NTBEA set up was the same with $k=1$ and $\epsilon=0.2$.
    
    Two final experiments are run comparing the best configurations obtained from the grid search. We are selecting the highest win ratio between all the configurations. This is not meant to be a \textit{fair} comparison between the algorithms. Within the possible agent's configurations we could probably find some equilibria without a strong dominance of an algorithm over the other, this however goes beyond the scope of this paper. The first experiment consists in playing \FPS~ while the second running a round robin tournament between the three.

\section{Experiments and Results}
        The preliminary experiment show an indistinguishable win ratio of the players, uniform random, each player 25\% (ignoring stalemates). Through this experiment we also tested the average duration of completely random games: (mean=140.86, sd=21.35, max=183, min=29), this helped us setting the timeout limit: 300, more than twice the average duration.
    
    \subsection{Grid Search}
        The values for the hyper-parameter space were all hand-picked and they can be seen on Table \ref{tab:hpspace}. This table also includes the best agents' configuration in bold, later referred to as BMRH*, SRH* and MCTS*, respectively.
        
\begin{figure}[!h]
            \centering
            \includegraphics[width=\columnwidth]{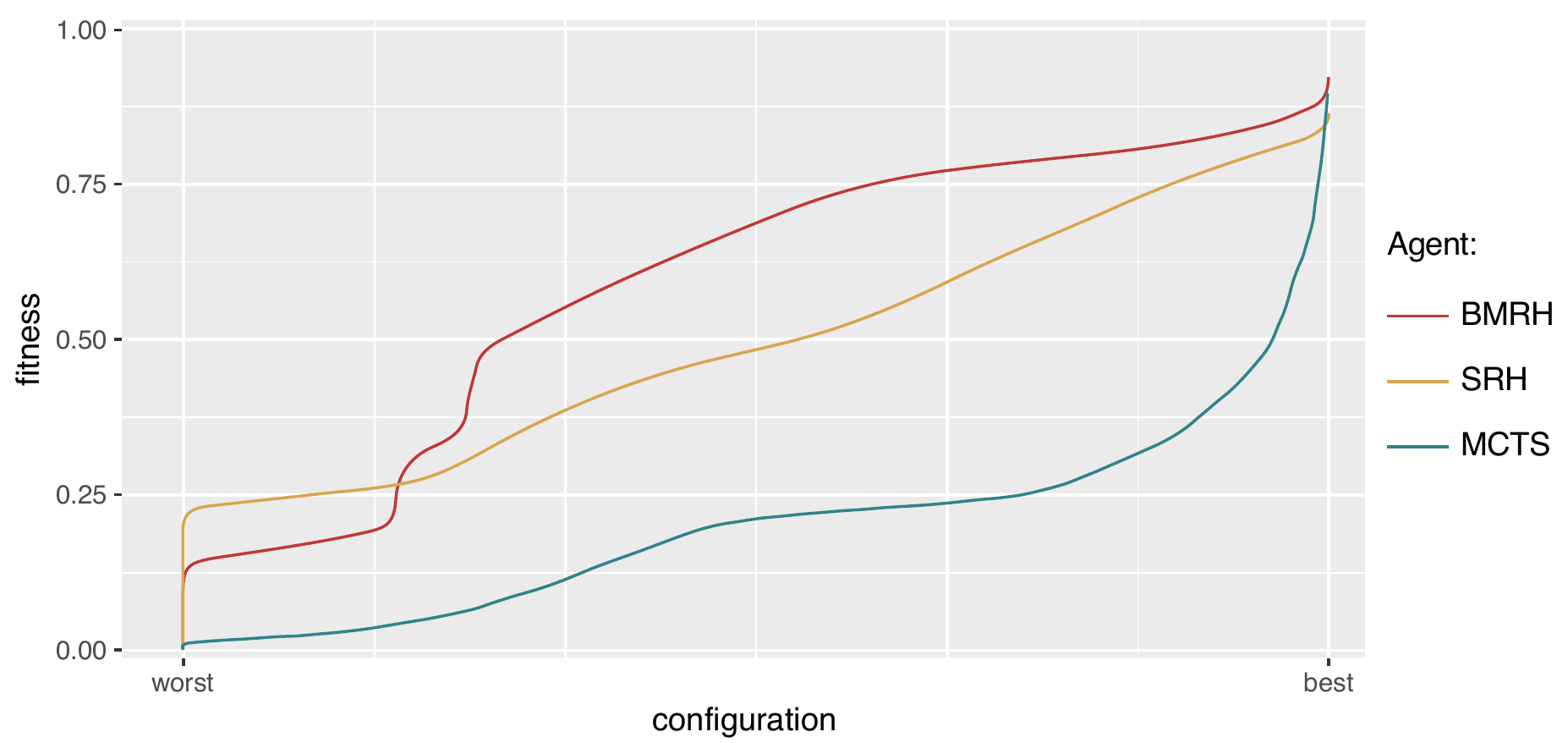}
            \caption{\label{fig:dist} hyper-parameter spaces' ordered fitness landscape.}
        \end{figure}

        Once the configurations were tested we have plotted the ordered hyper-parameter space, see Figure~\ref{fig:dist}. While analysing these plots we should keep in mind that the probability of winning a game out of luck is 0.25 (since it's a 4-player game) if all agents played uniformly at random. Figure \ref{fig:dist} shows in red BMRH's. We can notice how most configurations are concentrated in the higher half of the win rate making it a simpler agent to configure against OSLA. There are however a few configurations that perform below 0.25. SRH's, shown in yellow in Figure~\ref{fig:dist}, has a more robust behaviour (even with the poorest configurations) as we can see from the win ratio hardly going below 0.25, however its best configurations are performing slightly worse compared to BMRH's best ones. On the other hand, MCTS's hyper-parameter space shows how its configuration needs to be done carefully tuned to achieve a strong performance, being able to perform only with a handful configurations (see blue plot in Figure~\ref{fig:dist}).
        
        The best configurations of BMRH, SRH and MCTS scored respectively 0.924, 0.882 and 0.918.
        Looking at the parameters picked we can see that all the agents prefer estimating very short-term action plans. Both BMRH* and SRH* evolve sequences long just 2 actions and MCTS* grows a tree of maximum depth 2. This is probably due to the high stochasticity introduced by opponents' actions and random card shuffling. Keep re-sampling the short horizon is safer than adventuring in longer and dangerously uncertain plans. Along the same lines, the agents are always configured to not model the opponents. This is probably because having a weak model introduces even more noise reducing the overall budget significantly. A peculiarity of MCTS* is that UCB is tuned to completely eliminate the exploration term. This fundamentally means that whenever it is not expanding another action it is re-sampling the action with the highest expected reward (highest score) and wait for it to eventually drop because of re-sampling or expansion.
        BMRH* uses the branch mutation described earlier instead of a uniform random, however whether this brings better performance is not clear since the sequence evolved is only two-actions-long.
        
    \subsection{NTBEA}
    
\begin{figure}[!h]
            \centering
            \includegraphics[width=\columnwidth]{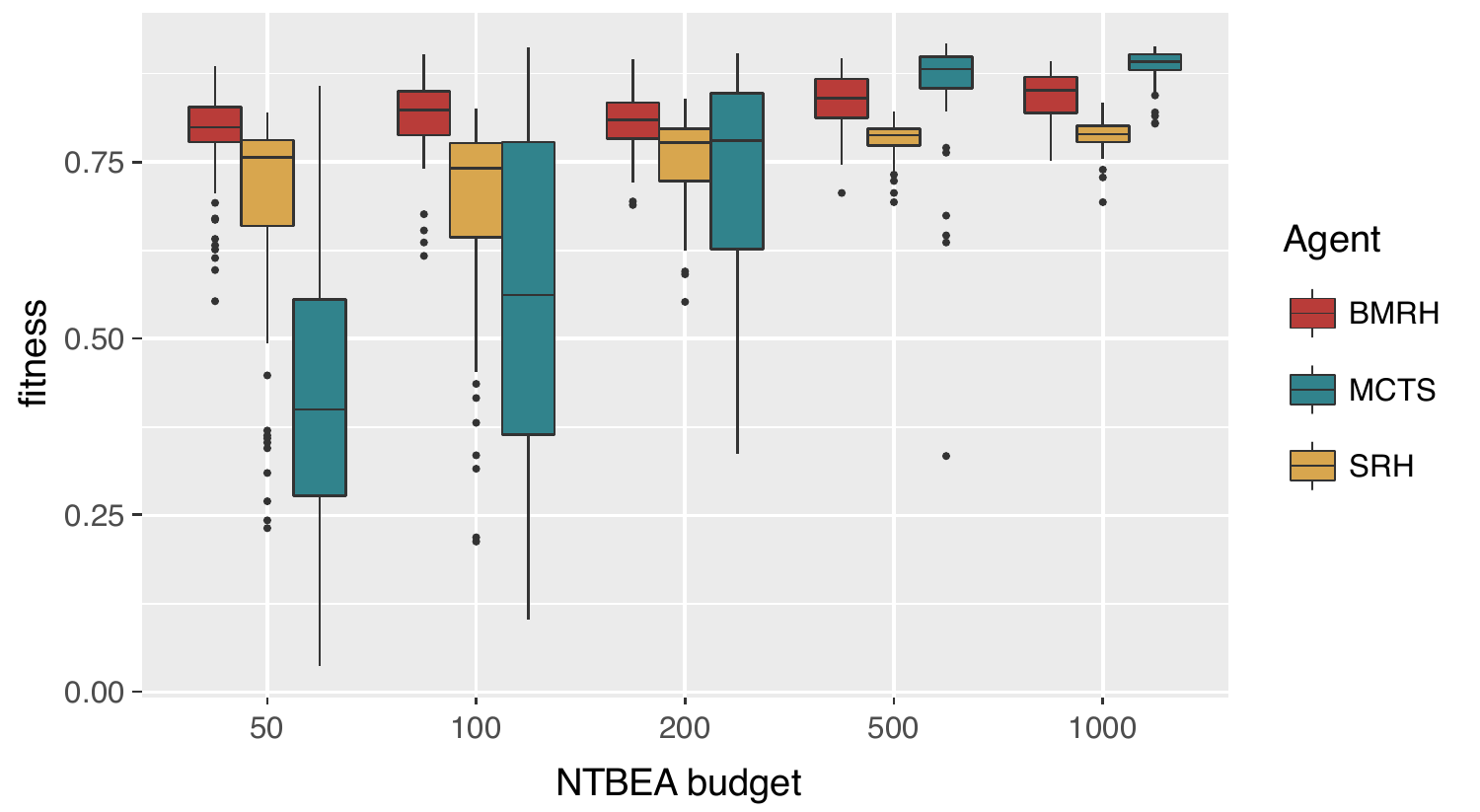}
            \caption{\label{fig:ntbea_comparison} Box-plots showing the NTBEA's outcome distributions varying budget and agent to optimise.}
        \end{figure}

        For each agent (BMRH, SRH and MCTS) we ran NTBEA with the following budgets: 50, 100, 200, 500 and 1000;
        We can see the all the results in Figure \ref{fig:ntbea_comparison}. It shows box-plots of the tuned agents' true fitness. Tuning MCTS is clearly harder with less budget than the other agents, but this is trivial looking at Figure \ref{fig:dist}. The important take-away from these experiments is that with only 1000 games played it's likely to get a good configuration of the agents. 

       \begin{table}[!h]
        \centering
        \caption{Round robin tournament results.}
        \begin{tabular}{l|c|c|c}
        \textbf{P1 vs P2}     & \textbf{P1 win rate} & \textbf{P2 win rate}  & \textbf{SM} \\
        \hline
        MCTS* vs BMRH*       &52.3\%                & 47.5\%                &0.2\%      \\
        SRH* vs BMRH*        &40.2\%                & 59.5\%                &0.3\%      \\
        SRH* vs MCTS*        &39.8\%                & 59.3\%                &1.6\%      
        \end{tabular}
    \label{par:rr}
    \end{table} 
    \subsection{Comparing Best Settings}
        Once obtained the best configurations of the agents we ran 10000 games of \FPS~between BMRH*, SRH*, MCTS* and an OSLA player as fourth player.
        This experiment doesn't aim to prove the general superiority of an algorithm over the other. It rather highlights the relative performance of agents that were separately tuned against weaker opponents.
        MCTS* and BMRH* have comparable performance (considering their std errors), respectively 35.67\%~($\pm$0.48\%) and 37.67\%~($\pm$0.48\%). SRH instead clearly has lower win ratio but it still manages to win around 25.09\%~($\pm$0.48\%) of the games. The bad performance shown by OSLA was expected, since it was the objective the agents were optimised for. Between the 10000 games, only the 1.39\% ended in a stalemate, well below the 14\% of completely random games. This proves that the agents have a clearer purpose in their strategy even with a simple heuristic.
        Finally we have run a round robin tournament on the two-player version with the following results, see Table \ref{par:rr} all the results are reported with a std error of $\pm$1.6\%. MCTS* can be slightly more robust than BMRH* in a 2 player-game where uncertainty due to number of opponents is lower. Generally both MCTS* and BMRH* outperform SRH*.

\section{Discussion and Future Work}
    In this paper we have presented a new framework for Game AI research. It presents big challenges due to its nature: multiplayer, stochastic and with a partially-observable state.
    This benchmark is efficient in its implementation, simulating 1.74 million states per second.
    The framework was tested on the \FPS~ version of the game without exploring variations of the game parameters, this limit was imposed to first assess the suitability of the agent to fast parameter-tuning.
    
    We introduced several baseline game-playing algorithms and shown how they can be efficiently tuned obtaining good performance in few game simulations even in a non-favourable hyper-parameter space. This feature makes the agents and the framework suitable to run experiments that require solid AI performance without a known testing condition e.g. when changing the game's parameters. The agents were provided with a very basic heuristic: player's prestige points. This poses a limit to the agent's skill potential, but it reduces the bias towards some game states thus it isn't a dramatic limitation for these initial experiments. In the future, when optimising against strong opponents it will likely be a crucial point.
    
    To fully take advantage of \game~the framework will need a PCG module to generate cards and noble tiles for configurations of the game were the number of token types varies. It would also allow to modify the starred parameters in Table \ref{par:setup}. Simulation-based PCG methods will highly benefit from the quick tunability of the agents introduced in this paper.
    
    Future work can be done to expand the game-playing agents available in the framework and to introduce more enhancements to the ones presented, both RHEA and MCTS are flexible methods.
    In real \gameS, predicting opponent's actions is key to competitive playing, so being able to use a reliable opponent model will critically improve the skill level of a player. That is a field that requires more attention and \gameR seems a perfect platform to expand the current state of the art.
 
\section*{Acknowledgements}
    \addcontentsline{toc}{section}{Acknowledgements}
    This work was funded by the EPSRC CDT in Intelligent Games and Game Intelligence (IGGI) EP/L015846/1.

\bibliographystyle{IEEEtran}
\bibliography{ms}

\end{document}